\documentclass[letterpaper, 10 pt, conference]{ieeeconf}

\IEEEoverridecommandlockouts                           
\usepackage{times}
\usepackage{graphics} % for pdf, bitmapped graphics files
\usepackage{mathtools}
\usepackage{graphicx}
\usepackage{tabularx}
\usepackage{caption}
\usepackage{wrapfig}
\usepackage{romannum}
\usepackage{amsmath,amssymb}
\usepackage{url}
\usepackage{bm}
\usepackage[table,xcdraw]{xcolor}
\usepackage{hyperref}
\usepackage{subfigure}
\usepackage{cite}
\usepackage{booktabs}

\usepackage{multirow}
\usepackage{booktabs}
\usepackage{pifont}
\usepackage{arydshln}
\usepackage{tabularx}
\usepackage[ruled,vlined]{algorithm2e}
\usepackage[noend]{algpseudocode}
\makeatletter
\let\OldStatex\Statex
\renewcommand{\Statex}[1][3]{%
  \setlength\@tempdima{\algorithmicindent}%
  \OldStatex\hskip\dimexpr#1\@tempdima\relax}
\renewcommand{\ALG@beginalgorithmic}{\normalsize}

\DeclareCaptionFont{mysize}{\fontsize{8}{9.6}\selectfont}
\title{\LARGE \bf  LAMP: Lane-Aligned Motion Primitives \\ for Feasible Trajectory Prediction}

\author{Sangjin Han$^{\dagger}$, Hoseong Jung$^{\dagger}$, Jeongtae Her$^{\dagger}$, Changhyun Choi, H. Jin Kim$^{*}$
\thanks{This work was partly supported by Institute of Information \& communications Technology Planning \& Evaluation (IITP) grant funded by the Korea government (MSIT) [NO.RS-2021-II211343, Artificial Intelligence Graduate School Program (Seoul National University)] and Hyundai Motor Chung Mong-Koo Foundation.}
\thanks{Sangjin Han, Hoseong Jung, Changhyun Choi and H. Jin Kim are with Seoul National University
{\tt\footnotesize \ (evertrue0504, ghtjdaleka, windust7, hjinkim@snu.ac.kr)}. 
Jeongtae Her is with the Hyundai Motor Company, Republic of Korea{\tt\footnotesize \ (jthuh@hyundai.com)}}
\thanks{$^*$ Corresponding author. $^\dagger$ These authors contributed equally to this work.}}%

\begin{document}

\maketitle

%%%%%%%%%%%%%%%%%%%%%%%%%%%%%%%%%%%%%%%%%%%%%%%%%%%%%%%%%%%%%%%%%%%%%%%%%%%%%%%%

\begin{abstract}
Motion forecasting is essential for autonomous driving systems to enable safe decision-making and planning in complex driving scenarios.
While existing predictors excel at minimizing standard displacement errors, they often overlook the adherence to lane topology of multimodal predictions, particularly for lower-probability modes.
Consequently, predicted trajectories may violate physical and logical constraints, making the prediction set unreliable for safety-critical planning.
In this paper, we propose LAMP (Lane-Aligned Motion Primitives), a topology-aware forecasting framework that anchors multimodal prediction to structured motion primitives aligned with lane topology.
Specifically, we use a VQ-VAE to learn shape-aware motion primitives as discrete intention queries, capturing spatiotemporal patterns beyond endpoint-based intentions.
We further introduce a feasibility-aware intention selector trained with a lane-topology prior for filtering unreachable intention queries, guiding the decoder to prioritize topology-consistent intentions while preserving behavioral diversity.
Extensive experiments on the Argoverse 2 dataset demonstrate that LAMP achieves prediction accuracy comparable to state-of-the-art baselines while outperforming them in feasibility and diversity metrics.
\end{abstract}

%%%%%%%%%%%%%%%%%%%%%%%%%%%%%%%%%%%%%%%%%%%%%%%%%%%%%%%%%%%%%%%%%%%%%%%%%%%%%%%%
\section{INTRODUCTION}
Motion forecasting is a core component of autonomous driving systems as it predicts the future trajectories of surrounding agents based on their historical states and the map context~\cite{shi2022motion, gao2020vectornet}.
Due to inherent behavioral uncertainty and complex inter-agent interactions, future motions are intrinsically stochastic rather than deterministic~\cite{lee2017desire, gupta2018social, salzmann2020trajectron}.
Consequently, effective predictors must generate multiple plausible trajectories to capture diverse driving intentions~\cite{shi2022motion}.
Such multimodal predictions enable downstream planning modules to evaluate potential risks and ensure safe decision-making in dynamic traffic scenarios~\cite{bansal2019chauffeurnet}.

Existing approaches commonly rely on predefined intention priors, including trajectory anchors~\cite{chai2020multipath, phan2020covernet} and target points~\cite{zhao2021tnt, gu2021densetnt} to capture diverse driving intentions.
While these priors facilitate multimodal coverage and stabilize training, they are typically designed independently of the specific scene context~\cite{chai2020multipath, phan2020covernet}.
As a result, their scene-agnostic design often fails to align with the underlying lane topology, producing structurally infeasible trajectories.
Moreover, prevailing training objectives prioritize the displacement accuracy of the most probable mode~\cite{shi2024mtr}.
Such objectives often overlook the feasibility of alternative modes.
Therefore, lower-probability trajectories are more prone to violating lane constraints (e.g., off-road predictions), reducing their reliability for downstream planning.
These limitations call for a topology-aware motion forecasting framework that enforces feasibility across all predicted modes while preserving multimodal diversity.

\begin{figure}[t!]
\centering
\includegraphics[width=\columnwidth]{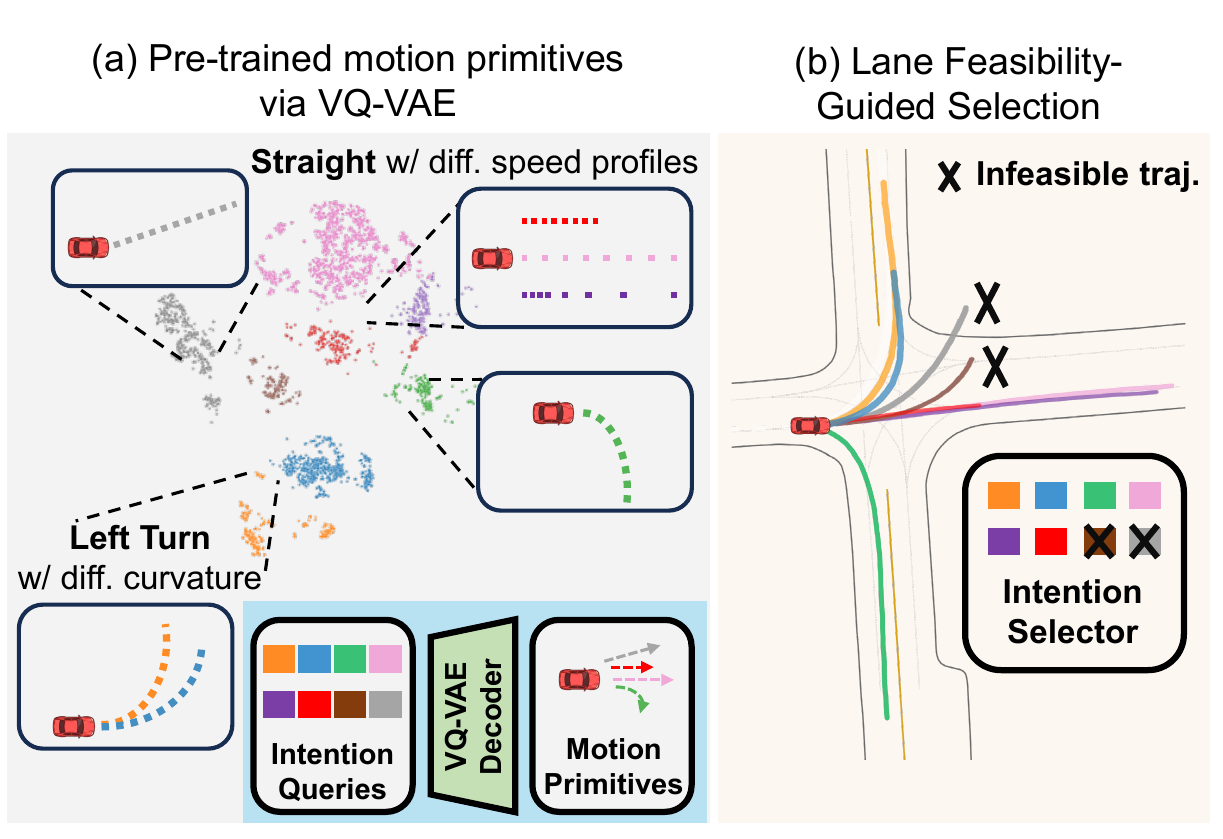}
\caption{\textbf{LAMP} improves the reliability of multimodal trajectory predictions. (a) A VQ-VAE provides a discrete set of intention queries; each query is decoded into a motion primitive representing a plausible future trajectory. (b) Infeasible intention queries are removed via lane-topology-guided selection, producing a feasible and diverse set of trajectories for downstream planning.
\vspace{-15pt}
}
\label{fig:main-figure}
\end{figure}

To address these challenges, we present \textbf{LAMP} (\textbf{L}ane-\textbf{A}ligned \textbf{M}otion \textbf{P}rimitives), which improves the reliability of multimodal prediction sets by combining structured motion primitives with topology-aware intention selection.
Rather than abstracting driving intentions as static endpoints, we model intention as a discrete set of learnable intention queries using a Vector Quantized Variational AutoEncoder (VQ-VAE)~\cite{van2017neural}.
Each query decodes into a trajectory prototype, which we refer to as a motion primitive that captures recurring driving behavior patterns as illustrated in Fig.~\ref{fig:main-figure}.
This formulation thereby preserves the rich spatiotemporal dynamics often lost in endpoint-based methods.
Beyond capturing dynamics, ensuring the alignment of motion primitives with the local road topology is essential to guarantee physical and logical feasibility.
Accordingly, we introduce a feasibility-aware intention selector that filters out off-road or unreachable queries prior to decoding.
This guides the model to generate a compact set of diverse trajectories that adhere to lane constraints for safety-critical planning.

We implement \textbf{LAMP} upon the Motion Transformer (MTR) architecture~\cite{shi2022motion}, within the UniTraj platform~\cite{feng2024unitraj}.
To evaluate both prediction accuracy and structural reliability, we report standard displacement metrics and further assess feasibility and diversity using established metrics and our refined variants.
Experiments on the Argoverse 2 dataset~\cite{benjamin2021argoverse2} demonstrate that \textbf{LAMP} achieves performance comparable with strong baselines on standard displacement metrics while substantially improving feasibility and diversity.

\begin{figure*}[h!]
\centering
\includegraphics[width=\textwidth]{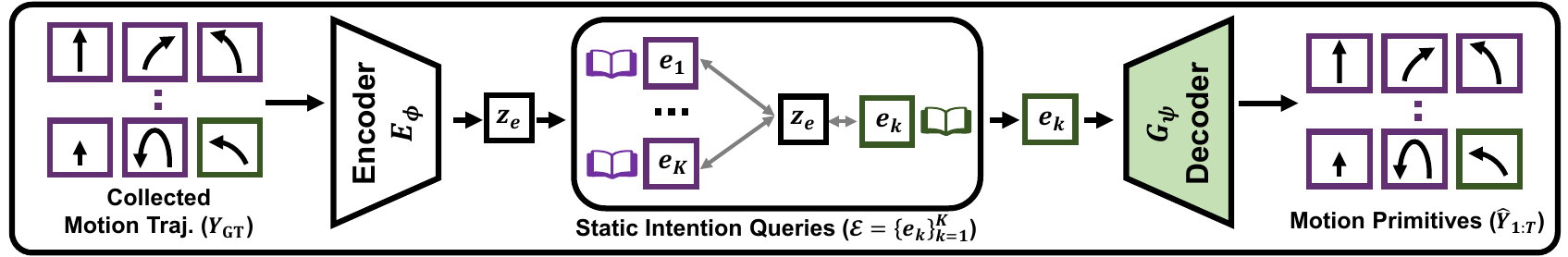}
\caption{Learning motion primitives with VQ-VAE~\cite{van2017neural}. Collected motion trajectories $Y_{\text{GT}}$ are encoded into a discrete codebook of intention queries to represent diverse driving intentions with finite codes.
Each code is decoded to a trajectory prototype $\hat{Y}_{1:T}$, which serves as a motion primitive in \textbf{LAMP}.
\vspace{-10pt}
}
\label{fig:VQ-VAE}
\end{figure*}

In summary, our contributions are threefold:
\begin{itemize}
    \item We propose \textbf{LAMP}, a topology-aware motion forecasting framework that improves the reliability of multimodal prediction sets via structured motion primitives and feasibility-aware intention selection.

    \item We formulate driving intention as discrete queries via a VQ-VAE, where each query decodes to a trajectory prototype capturing rich spatiotemporal dynamics.
    
    \item We introduce a lane-topology-guided intention selector that improves feasibility and diversity while preserving displacement accuracy.
\end{itemize}\label{sec:1}
%%%%%%%%%%%%%%%%%%%%%%%%%%%%%%%%%%%%%%%%%%%%%%%%%%%%%%%%%%%%%%%%%%%%%%%%%%%%%%%%

%%%%%%%%%%%%%%%%%%%%%%%%%%%%%%%%%%%%%%%%%%%%%%%%%%%%%%%%%%%%%%%%%%%%%%%%%%%%%%%%
\section{Related Work and Background}
\subsection{Modeling Multi-Modality in Motion Forecasting}\label{subsec:2.1}
Generating diverse and plausible future trajectories in motion forecasting is difficult due to the stochasticity of driving behavior.
Early latent generative methods sample multiple futures~\cite{lee2017desire, gupta2018social, salzmann2020trajectron, girgis2021latent}, but may produce collapsed or inconsistent modes.
Anchor-based approaches enhance multimodal coverage by relying on predefined intention priors such as trajectory templates~\cite{varadarajan2022multipath, phan2020covernet, zhao2021tnt, gu2021densetnt}, yet their scene-agnostic anchors may poorly align with complex lane topology.
Recent query-based methods~\cite{shi2022motion, ngiam2022scene, cheng2023forecast, nayakanti2023wayformer} employ learnable queries to interact with scene context.
This creates a trade-off: fixed queries can limit adaptability, while fully dynamic querying can be harder to optimize.
We address this trade-off using primitive-based intention queries and a lane-topology-guided intention selector for feasible and diverse multimodal forecasting.
\subsection{Intention Modeling for Reliable MultiModal Prediction}\label{subsec:2.2}
\subsubsection{Intention Representation}
Driving intention requires a representation that captures motion semantics beyond the endpoint.
Endpoint-based methods~\cite{zhao2021tnt, gu2021densetnt} model intention through discrete target candidates, but often miss trajectory shape and temporal evolution.
Instead, motion primitive-based representations encode driving intentions as trajectory prototypes.
Vector quantization models such as VQ-VAE~\cite{van2017neural} provide a discrete codebook of trajectory prototypes and have been explored for multimodal forecasting~\cite{zhang2024predicting, benaglia2024trajectory}, but can suffer from gradient or codebook collapse.
We adopt NSVQ~\cite{vali2022nsvq} to enable stable codebook learning from large-scale driving trajectories.

\subsubsection{Feasibility-Aware Selection}
Ensuring feasible multimodal predictions is critical for downstream planning~\cite{casas2020implicit}, as low-probability modes often violate map constraints~\cite{gao2020vectornet}.
Prior works typically improve feasibility through trajectory-level auxiliary losses~\cite{bansal2019chauffeurnet} or post-hoc heuristic corrections~\cite{zhao2021tnt}, but they do not explicitly select multimodal hypotheses based on their consistency with lane topology.
In contrast, we learn a lane-topology-guided intention selector that filters unreachable or off-road intention queries before decoding.
%%%%%%%%%%%%%%%%%%%%%%%%%%%%%%%%%%%%%%%%%%%%%%%%%%%%%%%%%%%%%%%%%%%%%%%%%%%%%%%%

%%%%%%%%%%%%%%%%%%%%%%%%%%%%%%%%%%%%%%%%%%%%%%%%%%%%%%%%%%%%%%%%%%%%%%%%%%%%%%%%
\section{METHODS}
\begin{figure*}[t!]
\centering
\includegraphics[width=\textwidth]{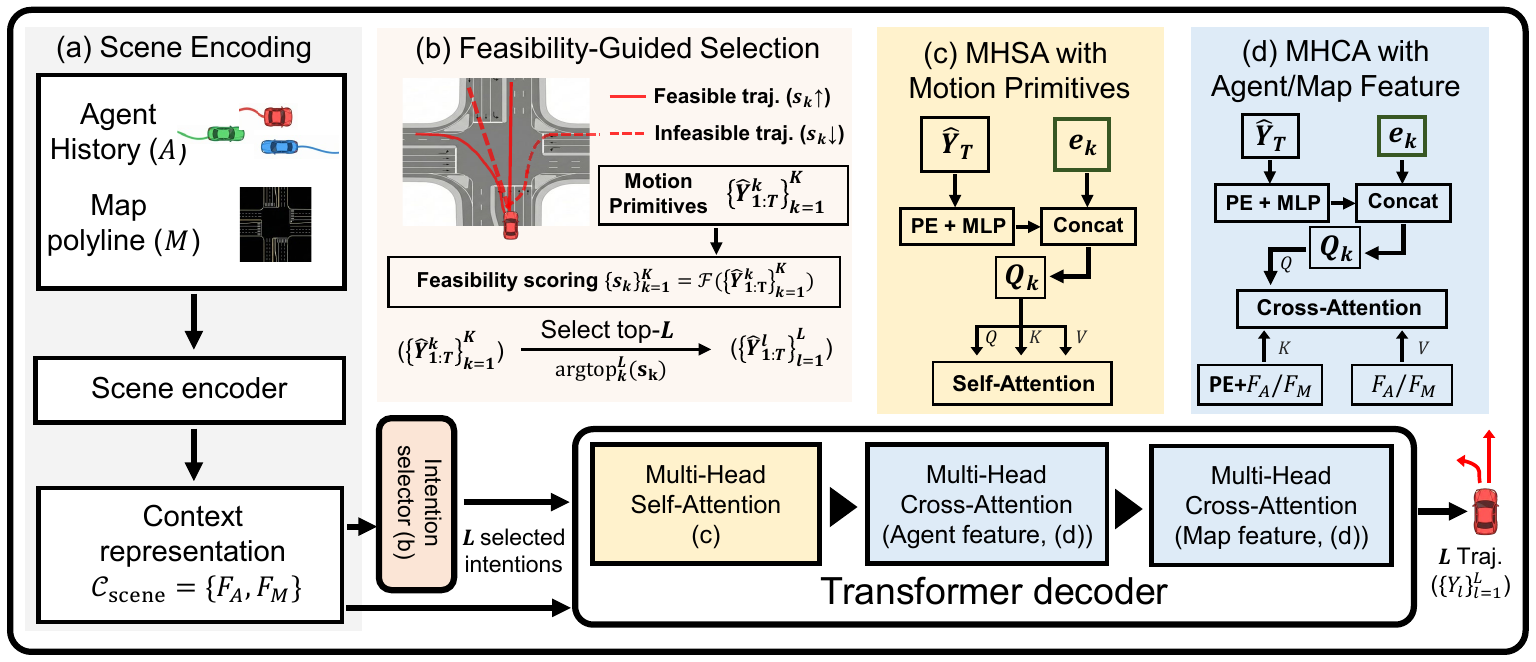}
\caption{Overview of the \textbf{LAMP} framework. (a) Scene context is encoded from agent histories and map polylines. (b) A feasibility-guided intention selector evaluates motion primitives and retains $L$ topology-consistent candidates from $K$ hypotheses. The selected intention embeddings and scene features are fed into a Transformer decoder. (c) Multi-head self-attention (MHSA) refines motion queries formed from static intention queries and motion primitives, followed by (d) cross-attention (MHCA) with agent and map context to generate $L$ trajectories.
\vspace{-10pt}
}
\label{fig:framework}
\end{figure*}

We propose \textbf{LAMP} (\textbf{L}ane-\textbf{A}ligned \textbf{M}otion \textbf{P}rimitives), a topology-aware motion forecasting framework that improves the reliability of multimodal prediction sets using structured motion primitives and feasibility-aware intention selection.
The overall framework is illustrated in Fig.~\ref{fig:framework}.
The remainder of this section details each component.
Sec.~\ref{subsec:3.1} describes the task formulation and the backbone architecture. 
Sec.~\ref{subsec:3.2} describes the learning of the discrete intention queries. 
Sec.~\ref{subsec:3.3} presents the lane-topology-guided intention selector for feasibility-aware selection.
Finally, Sec.~\ref{subsec:3.4} outlines the training and inference pipeline.
\subsection{Problem Formulation and Backbone Architecture}\label{subsec:3.1}
\subsubsection{Task Definition}
The goal of motion forecasting is to predict the future trajectory of a target agent over a horizon $T_f$, given historical context.
We adopt vectorized representations~\cite{gao2020vectornet}, where the input consists of agent histories $A\in\mathbb{R}^{N_a\times T_h \times C_a}$ and map polylines $M\in\mathbb{R}^{N_m\times N_p \times C_m}$.
Here, $N_a$, $T_h$, and $C_a$ denote the number of agents, past timesteps, and motion features, while $N_m$, $N_p$, and $C_m$ denote the number of polylines, points per polyline, and point attributes, respectively.
Following the agent-centric setting~\cite{varadarajan2022multipath}, inputs are normalized to the target's coordinate frame.
The model generates $L$ trajectories $\mathcal{Y}=\{Y_l, p_l\}_{l=1}^L$, where $Y_l\in\mathbb{R}^{T_f\times 2}$ is the predicted path and $p_l$ denotes the confidence.

\subsubsection{Scene Context Encoding}
We utilize the Motion Transformer (MTR)~\cite{shi2022motion} as the scene encoder.
Agent histories $A$ and map polylines $M$ are first embedded via PointNet-style MLPs to obtain initial features $F_{A}^{(0)}$ and $F_{M}^{(0)}$~\cite{qi2017pointnet}:
\begin{equation}
    F_{\text{A}}^{(0)} = \phi(\text{MLP}(A)), \quad F_{\text{M}}^{(0)} = \phi(\text{MLP}(M)),
\end{equation}
where $\phi$ denotes max-pooling.
To model local lane topology, MTR employs a local Transformer encoder that restricts attention to the $k$-nearest neighbors $\Omega(i)$:
\begin{equation}
    F^{(l+1)}[i] = \text{MHSA}\Big(h_i, \{h_j\}_{j\in\Omega(i)}\Big),
\end{equation}
where $h_i$ denotes the query token feature at the $l$-th layer, and $\{h_j\}$ represents the key and value features from the local neighborhood.
An auxiliary dense motion prediction (DMP) loss further enhances agent features:
\begin{equation}
    S_{\text{future}} = \text{MLP}(F_{\text{A}}), \quad \mathcal{L}_{\text{DMP}} = | S_{\text{future}} - S_{\text{GT}}|.
\end{equation}
This produces the final context representation $\mathcal{C}_\text{scene}=\{F_\text{A},F_\text{M}\}$ used in our subsequent decoding modules.

\subsubsection{Decoding Trajectories}
The standard MTR decoder employs a set of learnable queries to interact with the scene context via cross-attention.
Specifically, it defines $K$ static intention points $\mathcal{I}=\{I_k\}_{k=1}^K$ (e.g., via $k$-means clustering).
The intention query $Q_k$ is generated by encoding these points with positional embedding (PE):
\begin{equation}
    Q_k = \text{MLP}(\text{PE}(I_k)).
    \label{eq:MTR_query}
\end{equation}
These queries are then fed into the Transformer decoder to aggregate context features.
The final output is modeled as a Gaussian Mixture Model (GMM).
For a given time step $t$, the probability distribution of the agent's position $o$ is:
\begin{equation}
    P_t(o) = \sum_{k=1}^{K} p_k \cdot \mathcal{N}(o | \mu_k^t, \Sigma_k^t).
    \label{eq:MTR_GMM}
\end{equation}
The model is trained end-to-end by minimizing the negative log-likelihood loss for the best-matching mode (Top-1) combined with the auxiliary regression loss:
\begin{equation}
    \mathcal{L}_{\text{BASE}} = -\log P(\hat{Y}_{\text{GT}}) + \lambda_\text{DMP} \mathcal{L}_{\text{DMP}}.
    \label{eq:base_loss}
\end{equation}
In the following sections, we introduce our intention queries and feasibility-aware selection on top of this backbone.
\subsection{Learning Intention Queries and Motion Primitives}\label{subsec:3.2}
We learn a discrete codebook of intention queries using a NSVQ-based VQ-VAE~\cite{van2017neural, vali2022nsvq}, replacing the endpoint-based intention priors derived via $K$-means clustering in MTR~\cite{shi2022motion}.
Fig.~\ref{fig:VQ-VAE} illustrates the quantization pipeline.
Given a collected motion trajectory $Y_\text{GT}\in\mathbb{R}^{T\times 2}$, an encoder $E_\phi$ maps $Y_\text{GT}$ into a latent representation $z_e$.
We maintain a learnable codebook $\mathcal{E} = \{e_k\}_{k=1}^{K} \subset \mathbb{R}^{D_e}$, where $K$ is the codebook size and $D_e$ the embedding dimension.
The latent $z_e$ is quantized to the nearest code vector $z_q$:
\begin{equation}
    z_q = e_k \quad \text{where } k=\mathop{\text{argmin}}_{j} \|z_e - e_j\|_2^2. 
\end{equation}
A decoder $G_\psi$ reconstructs the trajectory prototype $\hat{Y}_{1:T}$ from $z_q$.
We adopt NSVQ~\cite{vali2022nsvq} to alleviate reliance on auxiliary commitment losses and to address the gradient-collapse issue in vector quantization.
Accordingly, we optimize the quantization module with the reconstruction objective:
\begin{equation}
    \mathcal{L}_{\text{VQ}} = \|Y_{\text{GT}} - \hat{Y}_{1:T}\|_2^2,
\end{equation}
where NSVQ provides nonzero gradients for both encoder outputs and codebook vectors during training.

This results in a discrete codebook $\mathcal{E}$ whose codes decode to trajectory prototypes (motion primitives) capturing diverse and realistic driving patterns.
For each code $e_k$, we form an intention query~$Q_k$ by combining the code embedding with a spatial grounding term.
Specifically, we extract the endpoint $\hat{Y}_T$ from its decoded trajectory prototype (via the frozen decoder $G_\psi$) and concatenate its positional encoding with the code embedding:
\begin{equation}
    Q_k = \text{Concat}\big(e_k, \; \text{MLP}(\text{PE}(\hat{Y}_T)) \big).
\end{equation}
The enriched queries $\mathcal{Q}=\{Q_k\}_{k=1}^K$ are then fed into the motion decoder to interact with the scene context.

\begin{table*}[t!]
\vspace{0.2cm}
\centering
\caption{Argoverse 2~\cite{benjamin2021argoverse2} validation results for main comparisons (Sec.~\ref{subsec:4.2}) and ablations (Sec.~\ref{subsec:4.3}). We report accuracy (b-minADE$_6$, b-minFDE$_6$, minADE$_6$, minFDE$_6$), feasibility (DAC, FR), and diversity (APD$_6$, FPD$_6$, DwF). Bold entries denote the best result for each metric within each block.}\label{tab:main_results}
\resizebox{0.9\textwidth}{!}{%
\begin{tabular}{c|l|cccc|cc|ccc}
\hline
& & \multicolumn{4}{c|}{Accuracy} & \multicolumn{2}{c|}{Feasibility} & \multicolumn{3}{c}{Diversity} \\
&Models & b-minADE$_6$$\downarrow$ & b-minFDE$_6$$\downarrow$ & minADE$_6$$\downarrow$ & minFDE$_6$$\downarrow$ & DAC$\uparrow$ & FR$\uparrow$ & APD$_6$ $\uparrow$ & FPD$_6$ $\uparrow$ & DwF$\uparrow$ \\ \hline
 \parbox[t]{2mm}{\multirow{6}{*}{\rotatebox[origin=c]{90}{\scalebox{0.8}{Main Results} }}} &\textbf{LAMP} (Ours) & 1.345 & 2.281 & 0.904 & 1.785 & \textbf{0.942} & \textbf{0.774} & \textbf{6.401} & \textbf{16.232} & \textbf{12.559} \\
&\textbf{MTR}~\cite{shi2022motion} & \textbf{1.315} & 2.127 & 0.842 & 1.667 & 0.925 & 0.665 & 4.109 & 10.963 & 7.811 \\
&\textbf{Wayformer}~\cite{nayakanti2023wayformer} & 1.420 & 2.281 & 0.804 & 1.664 & 0.905 & 0.691 & 4.157 & 11.151 & 8.173 \\
&\textbf{Forecast-MAE}~\cite{cheng2023forecast} & 1.379 & \textbf{2.125} & 0.751 & 1.492 & 0.937 & 0.692 & 2.983 & 7.728 & 5.657 \\
&\textbf{EMP}~\cite{prutsch2024efficient} & 1.385 & 2.141 & \textbf{0.749} & \textbf{1.503} & 0.934 & 0.684 & 2.936 & 7.608 & 5.524 \\
&\textbf{Autobot}~\cite{girgis2021latent} & 1.512 & 2.389 & 0.846 & 1.720 & 0.937 & 0.699 & 3.567 & 9.163 & 6.829 \\ \hline  \hline 

\parbox[t]{2mm}{\multirow{7}{*}
{\rotatebox[origin=c]{90}{\scalebox{0.8}{Ablation Study} }}} 
&\textbf{LAMP} (Ours) & 1.345 & \textbf{2.281} & \textbf{0.904} & \textbf{1.785} & 0.942 & 0.774 & 6.401 & 16.232 & \textbf{12.559} \\
& \textbf{VQ-VAE}~\cite{van2017neural} & 1.362 & 2.442 & 0.975 & 2.064 & 0.936 & 0.710 & 7.623 & 18.258 & 10.911 \\
& \textbf{FSQ}~\cite{mentzer2024finite} & 1.351 & 2.363 & 0.912 & 1.852 & \textbf{0.974} & \textbf{0.797} & 4.936 & 12.896 & 9.968 \\
& \textbf{NSVQ(128)} & 1.405 & 2.335 & 0.919 & 1.858 & 0.915 & 0.729 & 5.018 & 12.372 & 7.593 \\
& \textbf{NSVQ(32)} & 1.340 & 2.334 & 0.942 & 1.947 & 0.938 & 0.759 & 6.065 & 15.992 & 12.514  \\
& \textbf{w/o IS ($\mathbf{L=64}$)} & \textbf{1.339} & 2.313 & 0.923 & 1.911 & 0.898 & 0.713 & 5.447 & 14.521 & 10.961 \\ 
& \textbf{IS ($\mathbf{L=6}$)} & 1.426 & 2.655 & 1.065 & 2.302 & 0.875 & 0.758 & \textbf{8.981} & \textbf{18.404} & 11.461 \\
\hline
\end{tabular}%
}
\vspace{-10pt}
\end{table*}

\subsection{Feasibility-Aware Intention Selection}\label{subsec:3.3}
Given the $K$ motion primitives $\{\hat{Y}^k_{1:T}\}_{k=1}^K$ decoded from intention queries, we learn an intention selector that ranks and retains feasible modes before final decoding.
Passing all $K$ hypotheses to the Transformer decoder can mix infeasible intentions into the attention context, reducing the reliability of the final prediction set.
We therefore select only the top-$L$ feasible intentions and forward them to the decoder.
The selector is trained with a lane-topology prior $P_\text{lane}(k)$ and predicts a scene-conditioned categorical distribution over intentions, $P_{\text{pred}}(\;\cdot\;|\mathcal{C}_{\text{scene}})$.
We use $s_k = P_{\text{pred}}(k| \mathcal{C}_{\text{scene}})$ to select the top-$L$ intentions.

We define a feasibility scoring function $\mathcal{F}(\cdot)$ (Fig.~\ref{fig:framework}(b)) to provide a topology-aware supervision signal.
In this work, $\mathcal{F}(\cdot)$ is instantiated using a lane-topology prior.
Specifically, we automatically label a set of reachable lanes from the current agent pose by traversing HD-map lane connectivity.
We compute a distance $d_k$ between each hypothesis $\hat{Y}^k_{T}$ and the reachable-lane set and form a soft teacher distribution:
\begin{equation}
    P_{\text{lane}}(k)=\frac{\exp(-\alpha d_k)}{\sum_{j=1}^K \exp(-\alpha d_j)},
\end{equation}
where $\alpha$ controls the sharpness of the distribution.
The selector is trained to match the lane prior via:
\begin{equation}
    \mathcal{L}_\text{IS} = D_\text{KL}(P_\text{lane} \, \| \, P_{\text{pred}}).
\end{equation}
The selected intention embeddings are combined with scene features to form motion queries. 
These queries are refined by a Transformer decoder through multi-head self-attention and cross-attention (Fig.~\ref{fig:framework}(c,d)).
\subsection{Training and Inference Pipeline}\label{subsec:3.4}
\subsubsection{Multi-Stage Training Strategy}
Training proceeds in two stages.
In Stage 1, we train the VQ-VAE via $\mathcal{L}_\text{VQ}$ to learn a discrete codebook of intention queries and decoded motion primitives.
We freeze the VQ-VAE codebook and decoder and use the resulting intention queries as fixed prototypes.
In Stage 2, we jointly train the forecasting backbone and intention selector end-to-end by minimizing:  
\begin{equation}
    \mathcal{L}_{\text{TOTAL}} = \mathcal{L}_{\text{BASE}} + \lambda_\text{IS} \mathcal{L}_{\text{IS}}.
    \label{eq:loss_backbone}
\end{equation}

\subsubsection{Inference}
Our model generates $L$ trajectory hypotheses with confidence scores $\{p_k\}_{k=1}^{L}$.
For evaluation, we select $L_{\text{eval}}=6$ final predictions using non-maximum suppression (NMS).
Following MTR~\cite{shi2022motion}, we greedily pick the highest-confidence hypothesis and suppress other candidates whose endpoint distance to a selected one is below 2.5m, until $L_{\text{eval}}$ trajectories are obtained. All metrics are computed on this NMS-selected set unless otherwise specified.
%%%%%%%%%%%%%%%%%%%%%%%%%%%%%%%%%%%%%%%%%%%%%%%%%%%%%%%%%%%%%%%%%%%%%%%%%%%%%%%%

%%%%%%%%%%%%%%%%%%%%%%%%%%%%%%%%%%%%%%%%%%%%%%%%%%%%%%%%%%%%%%%%%%%%%%%%%%%%%%%%
\section{Experiments}
\subsection{Experimental Setup}\label{subsec:4.1}
We evaluate our framework on the Argoverse 2 Motion Forecasting Dataset~\cite{benjamin2021argoverse2}, which contains 250K scenarios with diverse road geometries and agent interactions.
Given 5 seconds of history ($T_h$), the task is to predict the target agent's future 6-second trajectory ($T_f$).
Our implementation follows the UniTraj~\cite{feng2024unitraj} framework with the official MTR~\cite{shi2022motion} backbone.
Training is performed on 4 NVIDIA RTX A5000 GPUs, while inference is measured on a single A5000 GPU with batch size 1; LAMP runs at 36.47 ms per scenario, comparable to MTR's 33.84 ms.
For intention query learning, we train an NSVQ-based VQ-VAE with $D_e{=}512$ and $K{=}64$ using Adam for 100 epochs, with a learning rate of $10^{-4}$ and batch size 256.
We train the forecasting model using AdamW for 32 epochs with the same learning rate and batch size 64, and set $L=16$, $\alpha=1.0$, and $\lambda_\text{IS}=50$.

We compare \textbf{LAMP} against strong baselines in the motion forecasting literature.
Specifically, we include \textbf{MTR}~\cite{shi2022motion}, \textbf{Wayformer}~\cite{nayakanti2023wayformer}, \textbf{Forecast-MAE}~\cite{cheng2023forecast}, \textbf{EMP}~\cite{prutsch2024efficient}, and \textbf{Autobot}~\cite{girgis2021latent}.
All results are reported on the validation set following the standard evaluation protocol.

We report standard displacement metrics for prediction accuracy (\textbf{b-minADE$_6$}, \textbf{b-minFDE$_6$}, \textbf{minADE$_6$} and \textbf{minFDE$_6$})~\cite{benjamin2021argoverse2, chen2024criteria}.
For structural reliability, we report two feasibility metrics with different strictness: \textbf{DAC}~\cite{park2020diverse} measures whether trajectories stay within the drivable area, while \textbf{FR} further checks traffic-rule constraints (e.g., lane connectivity).
For diversity, we include self-distance metrics used in prior work~\cite{yuan2020dlow, ma2021likelihood} (\textbf{APD$_6$}, \textbf{FPD$_6$}), and introduce \textbf{DwF} (Diversity-while-Feasible), a feasibility-aware diversity metric.
DwF measures diversity by averaging pairwise endpoint distances over the feasible subset.
\subsection{Main Results}\label{subsec:4.2}
\begin{figure}[t!]
\centering
\includegraphics[width=\columnwidth]{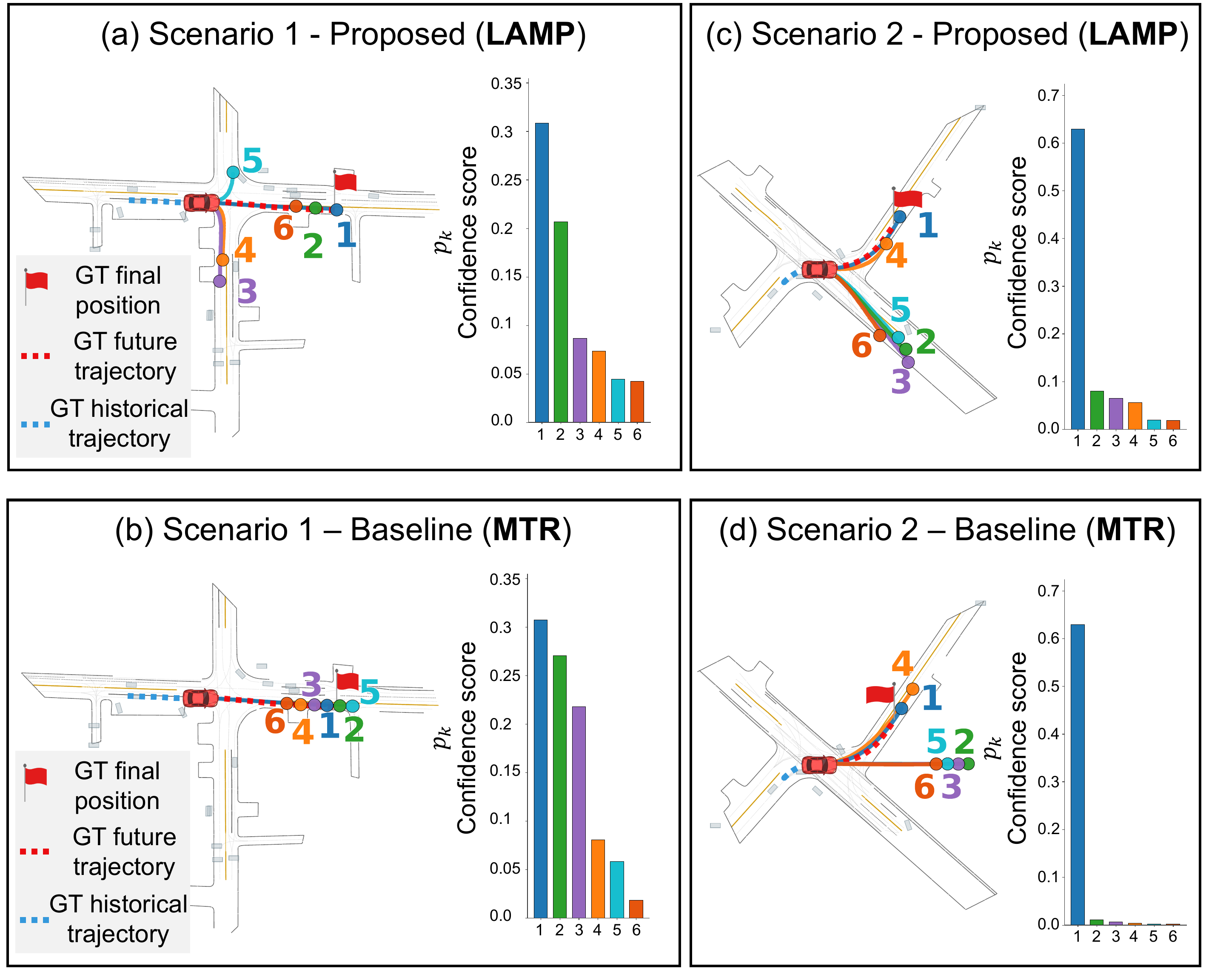}
\caption{Qualitative comparison of \textbf{LAMP} and \textbf{MTR} on two intersection scenarios.
Each panel visualizes multimodal predictions with confidence scores $p_k$, along with agent history, ground-truth future, and final position.
Scenario 1 highlights improved diversity of \textbf{LAMP}, while Scenario 2 shows improved feasibility by reducing off-road predictions compared to \textbf{MTR}.
}
\vspace{-10pt}
\label{fig:qual-results}
\end{figure}

Table~\ref{tab:main_results} presents the quantitative results on Argoverse 2.
\textbf{LAMP} substantially improves the feasibility and diversity of multimodal prediction sets while maintaining competitive prediction accuracy.
Compared to strong baselines, our gains are most pronounced in feasibility under stricter constraints.
In particular, DAC primarily captures whether a trajectory remains within the drivable area, whereas FR further evaluates lane topology and traffic-rule consistency.
Since downstream planners typically reject hypotheses that are unreachable or rule-violating  even when they are nominally on-road, improvements in FR are more indicative of planner-relevant feasibility.
Accordingly, \textbf{LAMP} exhibits a larger improvement in FR than in DAC, indicating that \textbf{LAMP} reduces not only off-road trajectories but also traffic rule-violating motions.
\textbf{LAMP} also improves diversity in a planner-relevant manner.
For downstream planning, diversity is useful only when the multimodal set remains feasible.
Thanks to the learned motion primitives from NSVQ-based VQ-VAE, \textbf{LAMP} increases overall diversity (APD$_6$/FPD$_6$), indicating improved intention coverage.
This combination provides a reliable multimodal prediction set that offers diverse yet feasible hypotheses for downstream planners.
\subsection{Ablation Study}\label{subsec:4.3}
We conduct ablations on the quantization method, NSVQ codebook size, and lane-topology-guided intention selector.
Under the same backbone and training protocol, NSVQ outperforms VQ-VAE~\cite{van2017neural} and FSQ~\cite{mentzer2024finite}, likely due to more stable optimization and better codebook utilization, therefore is adopted in our final model.
For the codebook size, NSVQ with $K{=}64$ achieves the best trade-off.
NSVQ(32) lacks limited codebook capacity, which restricts the expressiveness of intention queries, whereas NSVQ(128) suffers from degraded performance due to codebook collapse caused by imbalanced updates under fixed top-$L$ selection.
Finally, ablating the intention selector shows that our full model, IS($L{=}16$), substantially improves feasibility and diversity over w/o IS($L{=}64$), confirming that selecting planner-relevant feasible intentions benefits the Transformer decoder.
In contrast, IS($L{=}6$) reduces feasibility and DwF despite increasing endpoint spread, suggesting that too few intentions limit decoder capacity and weaken trajectory refinement.
\subsection{Qualitative Analysis}\label{subsec:4.4}
Fig.~\ref{fig:main-figure} visualizes the learned intention space and feasibility-aware selection in \textbf{LAMP}. 
The t-SNE projection (Fig.~\ref{fig:main-figure}a) shows that decoded intention queries exhibit interpretable driving patterns with varying curvatures and speeds. 
This suggests that the codebook effectively captures diverse trajectory shapes beyond simple endpoint-based priors. 
Subsequently, the intention selector filters out structurally infeasible primitives (Fig.~\ref{fig:main-figure}b), enhancing the reliability of the prediction set. 
Fig.~\ref{fig:qual-results} further demonstrates \textbf{LAMP}'s qualitative advantages over \textbf{MTR} on intersection scenarios. 
While \textbf{MTR} often concentrates on a single dominant mode (Scenario 1) or assigns confidence to off-road trajectories (Scenario 2), \textbf{LAMP} generates diverse, plausible maneuvers that strictly adhere to the lane topology. 
These results validate that our approach improves structural reliability while preserving multimodal coverage.
\subsection{Trade-off Analysis: Map-Adaptive Decoding via LoRA}
\begin{table}[t!]
\centering
\caption{Comparison of \textbf{LAMP} and \textbf{LAMP+LoRA} on the Argoverse~2 validation set. We report feasibility and diversity metrics.}
\label{tab:LoRA}
\begin{tabular}{l|ccccc}
\toprule
Model & DAC$\uparrow$ & FR$\uparrow$ & APD$_6$ $\uparrow$ & FPD$_6$ $\uparrow$ & DwF$\uparrow$ \\
\midrule
\textbf{LAMP}        & 0.942 & 0.774 & \textbf{6.401} & \textbf{16.232} & \textbf{12.559} \\
\textbf{LAMP + LoRA} & \textbf{0.983} & \textbf{0.794} & 3.943 & 10.402 & 8.092 \\
\bottomrule
\end{tabular}
\end{table}

\begin{figure}[t!]
\centering
\includegraphics[width=\columnwidth]{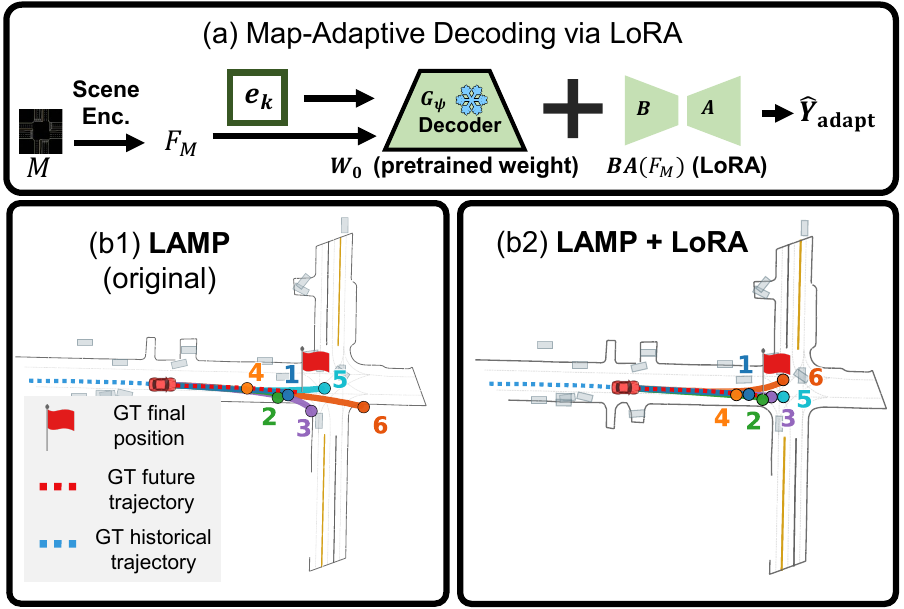}
\caption{LoRA-based map conditioning for motion-primitive reconstruction. (a) We apply LoRA~\cite{hu2022lora} to the NSVQ decoder to incorporate map features. (b) \textbf{LAMP+LoRA} increases feasibility but yields less diverse multimodal predictions than \textbf{LAMP}.
}
\vspace{-15pt}
\label{fig:LoRA}
\end{figure}

While \textbf{LAMP} learns scene-agnostic intention queries and decodes them into motion primitives, we additionally explored injecting map priors into the NSVQ decoder via Low-Rank Adaptation (LoRA)~\cite{hu2022lora}.
As illustrated in Fig.~\ref{fig:LoRA}(a), we freeze the pre-trained decoder weights $W_0$ and inject a lightweight low-rank update modulated by map features $F_{\text{M}}$:
\begin{equation}
h = W_0 x + \Delta W(F_{\text{M}})x = W_0 x + B A(F_{\text{M}}) x,
\end{equation}
which yields map-adaptive motion primitives $\hat{Y}^{k}_{\text{adapt}}$ during reconstruction.
Table~\ref{tab:LoRA} shows that map-adaptive variant (\textbf{LAMP+LoRA}) improves feasibility metrics, while it reduces diversity metrics.
Fig.~\ref{fig:LoRA}(b) further confirms this trend: compared to \textbf{LAMP}, \textbf{LAMP+LoRA} produces more lane-consistent predictions but a more concentrated set of hypotheses.
As shown in Fig.~\ref{fig:LoRA}, early map conditioning induces mode contraction; intention queries become heavily biased toward dominant lane-following behaviors, collapsing multimodal hypotheses. 
Due to this feasibility--diversity trade-off, we exclude LoRA from the final \textbf{LAMP} model, leaving diversity-preserving scene adaptation as future work.
%%%%%%%%%%%%%%%%%%%%%%%%%%%%%%%%%%%%%%%%%%%%%%%%%%%%%%%%%%%%%%%%%%%%%%%%%%%%%%%%

%%%%%%%%%%%%%%%%%%%%%%%%%%%%%%%%%%%%%%%%%%%%%%%%%%%%%%%%%%%%%%%%%%%%%%%%%%%%%%%%
\section{CONCLUSION}
In this paper, we introduced \textbf{LAMP}, a topology-aware framework for reliable multimodal motion forecasting.
\textbf{LAMP} learns reusable motion primitives through an NSVQ-based VQ-VAE and employs a lane-topology-guided intention selector to filter infeasible hypotheses before decoding.
This design enables diverse yet lane-consistent predictions, improving feasibility without sacrificing displacement accuracy on Argoverse 2.
Future work will integrate \textbf{LAMP} with downstream planning in closed-loop systems and evaluate its impact in real-world driving scenarios.
%%%%%%%%%%%%%%%%%%%%%%%%%%%%%%%%%%%%%%%%%%%%%%%%%%%%%%%%%%%%%%%%%%%%%%%%%%%%%%%%

\addtolength{\textheight}{0cm}   % This command serves to balance the column lengths

\bibliographystyle{IEEEtran}
\bibliographystyle{plainnat}

\end{document}